\documentclass[10pt,twocolumn,letterpaper]{article}

\usepackage{iccv}
\usepackage{times}
\usepackage{epsfig}
\usepackage{graphicx}
\usepackage{amsmath}
\usepackage{amssymb}
\usepackage{multirow}
\usepackage{booktabs}

\usepackage[pagebackref=true,breaklinks=true,letterpaper=true,colorlinks,bookmarks=false]{hyperref}

\iccvfinalcopy 

\def\iccvPaperID{5549} 

\ificcvfinal\fi
\newcommand{\figtext}[1]{{\footnotesize #1}}
\newcommand{\delete}[1]{{\color{red} }} 
\newcommand{\blue}[1]{#1} 

\begin{document}

\title{D3G: Exploring Gaussian Prior for Temporal Sentence Grounding with \\ Glance Annotation}

\author{Hanjun Li\textsuperscript{1},
Xiujun Shu\textsuperscript{1},
Sunan He\textsuperscript{2},
Ruizhi Qiao\textsuperscript{1},
Wei Wen\textsuperscript{1},
Taian Guo\textsuperscript{1},
Bei Gan\textsuperscript{1},
Xing Sun\textsuperscript{1}\thanks{Corresponding author}
\\
\textsuperscript{1}{Youtu Lab, Tencent} \quad 
\textsuperscript{2}{Hong Kong University of Science and Technology} 
\\
{\tt\small \{hanjunli, xiujunshu, ruizhiqiao, jawnrwen, taianguo, stylegan, winfredsun\}@tencent.com} \\
{\tt\small sunan.he@connect.ust.hk}
}

\maketitle
\ificcvfinal\thispagestyle{empty}\fi

\begin{abstract}
Temporal sentence grounding (TSG) aims to locate a specific moment from an untrimmed video with a given natural language query. Recently, weakly supervised methods still have a large performance gap compared to fully supervised ones, while the latter requires laborious timestamp annotations. In this study, \blue{we aim to reduce the annotation cost yet keep competitive performance for TSG task compared to fully supervised ones. To achieve this goal,} we investigate a recently proposed glance-supervised temporal sentence grounding task, which requires only single frame annotation (referred to as glance annotation) for each query. \delete{ and results in trivial annotation cost compared to weakly supervised counterparts.} Under this setup, we propose a \textbf{D}ynamic \textbf{G}aussian prior based \textbf{G}rounding framework with \textbf{G}lance annotation (D3G), which consists of a Semantic Alignment Group Contrastive Learning module (SA-GCL) and a Dynamic Gaussian prior Adjustment module (DGA). Specifically, SA-GCL samples reliable positive moments from a 2D temporal map via jointly leveraging Gaussian prior and semantic consistency, which contributes to aligning the positive sentence-moment pairs in the joint embedding space. Moreover, to alleviate the annotation bias resulting from glance annotation and model complex queries consisting of multiple events, we propose the DGA module, which adjusts the distribution dynamically to approximate the ground truth of target moments. Extensive experiments on three challenging benchmarks verify the effectiveness of the proposed D3G. It outperforms the state-of-the-art weakly supervised methods by a large margin and narrows the performance gap compared to fully supervised methods. Code is available at \url{https://github.com/solicucu/D3G}.

\end{abstract}

\begin{figure}
    \centering
    \includegraphics[width=\linewidth]{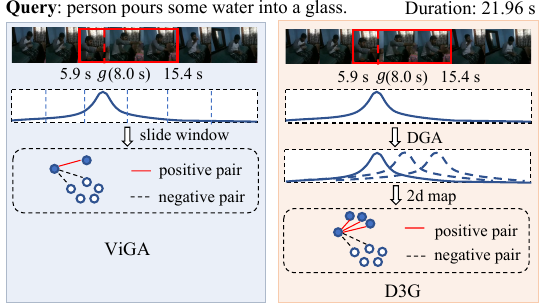}
    \caption{Illustration of glance annotation $g$ (red dashed line) and simple comparison between ViGA and D3G. The red rectangle indicates the boundary of target moment.}
    \label{fig:viga_vs_d3g}
    \vspace{-0.6cm}
\end{figure}

\section{Introduction}
Temporal sentence grounding is a fundamental problem in computer vision and receives an increasing attention in recent years.
Given the query sentence and an untrimmed video, the goal of TSG is to localize the start and end timestamps of specific moment that semantically corresponds to the query. In recent years, full supervised temporal sentence grounding (FS-TSG) has achieved tremendous achievements~\cite{gao2017tall,anne2017localizing,zhang2019man,zhang2020learning,xiao2021boundary,wang2021structured,wang2022negative, zhang2021multi}. However, obtaining accurate timestamps for each sentence is labor-intensive and subjective, which prevents it from scaling to large-scale video-sentence pairs and practical applications. 

Weakly supervised temporal sentence grounding (WS-TSG), which requires only the video and query pairs, receives an increasing attention recently. Although great advances~\cite{mithun2019weakly,wang2021fine,huang2021cross,zheng2022weakly, zhang2020learning, zheng2022weakly1} have been achieved in recent years, there still remains a huge performance gap between WS-TSG and FS-TSG. WS-TSG suffers from severe localization issues due to the large discrepancy between video-level annotations and clip-level task.  

Recently, Cui~\etal~\cite{cui2022video} propose a new annotating paradigm called glance annotation for TSG, requiring the timestamp of only random single frame within the temporal boundary of the target moment. It is noted that such annotation only increases trivial annotating cost compared to WS-TSG. \delete{This is because the annotators need to glance at specific frame and then write down the moment description. In this way, the glance annotation is obtained naturally.}
Figure~\ref{fig:viga_vs_d3g} illustrates the details of glance annotation. With glance annotation, Cui~\etal propose the ViGA based on contrastive learning. ViGA first cuts the input video into clips of fixed length, which are assigned with Gaussian weights generated according to the glance annotation, and contrasts clips with queries. There are two obvious disadvantages in this way. First, moments of interest usually have various durations. Therefore, these clips cannot cover a wide range of target moments, which inevitably aligns the sentence with incomplete moment and obtains sub-optimal performance. Second, ViGA utilizes a fixed scale Gaussian distribution centered at the glance frame to describe the span of each annotated moment. However, the glance annotations are not guaranteed at the center of target moments, which results in annotation bias as shown in Figure~\ref{fig:anno_bias}. 
Besides, since some complex query sentences consist of multiple events, a single Gaussian distribution is hard to cover all events at the same time as shown in Figure~\ref{fig:multi_events}. To address the aforementioned defects and fully unleash the potential of Gaussian prior knowledge with the low-cost glance annotation, we propose a \textbf{D}ynamic \textbf{G}aussian prior based \textbf{G}rounding framework with \textbf{G}lance annotation (D3G) as shown in Figure~\ref{fig:framework}. 

We first generate a wide range of candidate moments following 2D-TAN~\cite{zhang2020learning}. Afterwards, we propose a Semantic Alignment Group Contrastive Learning module (SA-GCL) to align the positive sentence-moment pairs in the joint embedding space.
Specifically, for each query sentence, we sample a group of positive moments \blue{according to calibrated Gausssian prior} and minimize the distances between these moments and the query sentence. In this way, it tends to gradually mine the moments which have increasing overlap with the ground truth. Moreover, we propose a Dynamic Gaussian prior Adjustment module (DGA), which further alleviates annotation bias and approximates the span of complex moments consisting of multiple events. 
Specifically, we adopt multiple Gaussian distributions to describe the weight distributions of moments. Therefore, the weight distributions for various moments can be flexibly adjusted and gradually approach to the ground truth. 
Our contributions are summarized as follows:

\begin{figure}[ht]
    \centering
    \includegraphics[width=0.7\linewidth]{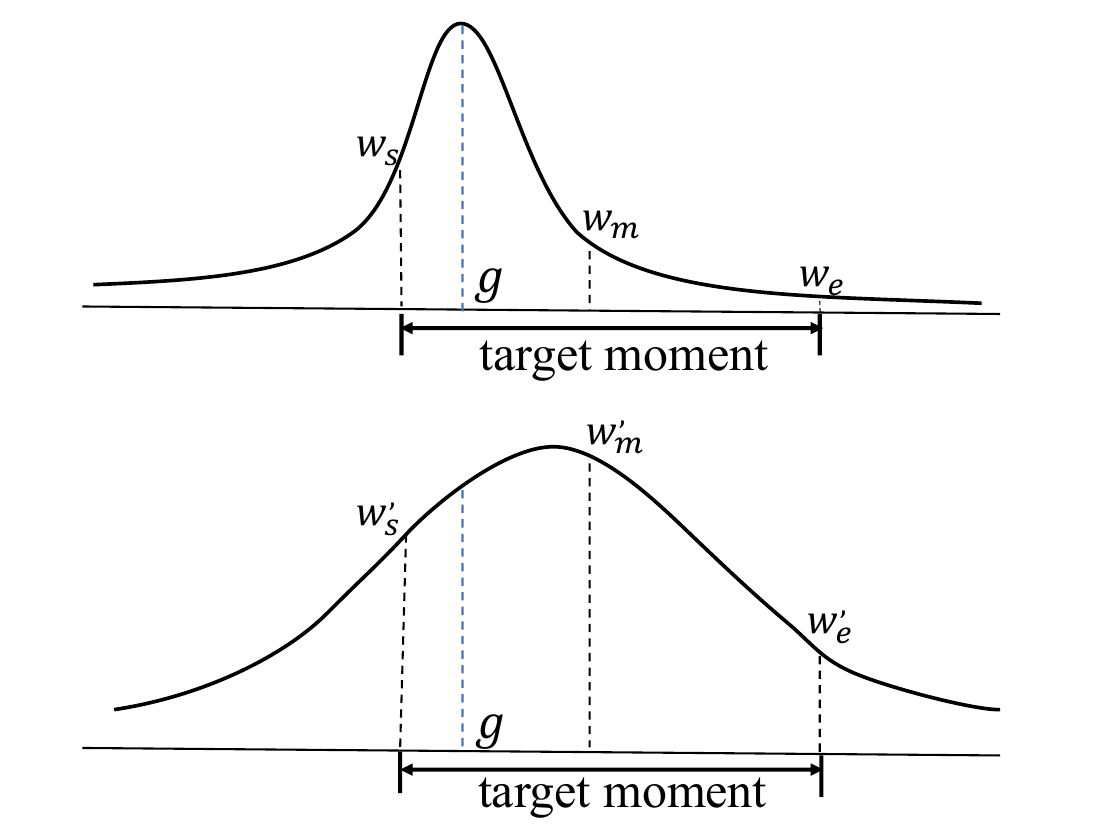}
    \caption{Illustration of annotation bias and Gaussian prior after dynamic adjustment. Top: the target moment is assigned with a low weight $w_m$ due to the bias of glance annotation according to ViGA, which we call annotation bias. Bottom: a reasonable Gaussian distribution is obtained via DGA described in Section~\ref{sec:dga}.}
    \label{fig:anno_bias}
\end{figure} 
\begin{itemize}
    \item 
    We propose a Dynamic Gaussian prior based Grounding framework with Glance annotation (D3G), which facilitates the development of temporal sentence grounding with lower annotated cost. 
    \item We propose a Semantic Alignment Group Contrastive Learning module to align the features of the positive sentence-moment pairs and a Dynamic Gaussian prior Adjustment module to ease the annotation bias and model the distributions of complex moments.
    \item Extensive experiments demonstrate that D3G obtains consistent and significant gains compared to method under the same annotating paradigm and outperforms weakly supervised methods by a large margin. 
\end{itemize}

\section{Related Work}
\noindent \textbf{Full Supervised Temporal Sentence Grounding.}
The FS-TSG methods can be categorized into two groups. 
Two-stage methods~\cite{anne2017localizing, gao2017tall, ge2019mac, jiang2019cross, jiao2018three, xu2019multilevel} first propose candidate segments in a video through sliding window or proposal generation. 
A cross-modal matching network is then employed to find the best matching clip.
However, these \textit{propose-and-match} paradigms are 
time-consuming due to the numerous candidates.
To reduce the redundant computation, some researchers proposed single-stage methods~\cite{chen2018temporally, chen2019localizing, yuan2019find, zhang2019man, wang2020temporally, mun2020local, zhang2020learning, nan2021interventional, yang2022video, yang2022tubedetr}.
2D-TAN~\cite{zhang2020learning} constructs 2D feature map to model the temporal relations of video segment.
Recently, Wang~\etal~\cite{wang2022negative} propose a Mutual Matching Network based on 2D-TAN, and further improve the performance via exploiting both intra- and inter-video negative samples.  
Although fully supervised methods achieve satisfying performance, they are highly dependent on accurate timestamp annotations. It is highly time-consuming and laborious to obtain these annotations for large-scale video-sentence pairs.
\par
\noindent \textbf{Weakly Supervised Temporal Sentence Grounding.}
Specifically, WS-TSG methods can be grouped into reconstruction-based methods~\cite{duan2018weakly, lin2020weakly, song2020weakly, chen2021towards} and multi-instance learning (MIL) methods~\cite{mithun2019weakly, gao2019wslln, chen2020look, yang2021local, huang2021cross, tan2021logan}.
SCN~\cite{lin2020weakly} employs a semantic completion network to recover the masked words in the query sentence with the generated proposals, which provides feedback for facilitating final predictions.
To further exploit the negative samples in MIL-based methods, CNM~\cite{zheng2022weakly1} and CPL~\cite{zheng2022weakly} propose to generate proposals with Gaussian functions and introduce intra-video contrastive learning.
WS-TSG methods indeed advance \delete{progressively} with low annotation cost, however, there still remains a large performance gap compared to FS-TSG methods due to the discrepancy between video-level annotations and clip-level task.
\par
\noindent \textbf{Glance Supervised Temporal Sentence Grounding.}
Recently, ViGA~\cite{cui2022video} proposes glance supervised TSG (GS-TSG) task with a new annotating paradigm.
ViGA utilizes a Gaussian function to model the relevance of different clips with target moment and contrasts the clips with the queries.
Though ViGA achieves promising performance, it still suffers from two limitations as mentioned in Introduction.
Concurrently, Xu~\etal~\cite{xu2022point} propose the similar task called PS-VTG, and generate pseudo segment-level labels based on language activation sequences. 
To better explore the Gaussian prior for TSG task with glance annotation, we propose a simple yet effective D3G, which achieves competitive performance compared with both WS-TSG and FS-TSG methods. Concurrent with our work, Ju~\etal~\cite{ju2023constraint} propose a robust partial-full union framework (PFU) and achieve excellent performance with glance annotation or short-clip labels.

\section{Proposed Method}
\subsection{Overview} 
Given an untrimmed video $V$ and query sentence $S$, the temporal sentence grounding task aims to determine the start timestamp $t_s$ and end timestamp $t_e$, where the moment $V_{t_s:t_e}$ best semantically corresponds to the query. As for FS-TSG, the exact timestamps ($t_s$,$t_e$) of corresponding moment is provided given a query description. In contrast, Cui~\etal~\cite{cui2022video} propose a new low-cost annotating paradigm called glance annotation, which requires only single timestamp $g$, satisfying $g \in [t_s, t_e]$.
Following the setting of ~\cite{cui2022video}, we propose a \textbf{D}ynamic \textbf{G}aussian prior based \textbf{G}rounding framework with \textbf{G}lance annotation (D3G) to fully unleash the potential of glance annotations.
\begin{figure}
    \centering
    \includegraphics[width=\linewidth]{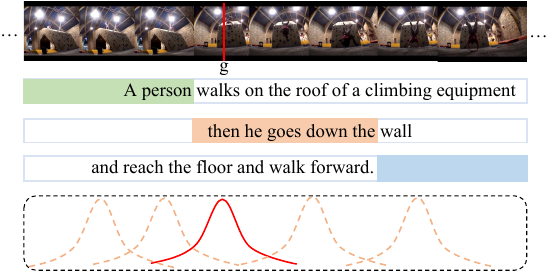}
    \caption{Illustration of complex query consists of multiple events. Note that $g$ indicates the position of glance annotation. The according Gaussian distribution (red curves) is hard to cover the whole target moments. We utilize DGA module to mine multiple latent Gaussian distributions (dashed line) to model such query. }
    \label{fig:multi_events}
\end{figure}

Our D3G adopts the network architecture similar to ~\cite{zhang2020learning,wang2022negative}.
Given an untrimmed video, we firstly encode the video into feature vectors with pre-trained 2D or 3D convolutional network~\cite{simonyan2014very,tran2015learning} and 
segment the video features into $N$ video clips. Specifically, we apply average pooling to each clip to obtain clip-level features $V = \{ f_{1}^{v}, f_{2}^{v}, ... , f_{N}^{v} \} \in \mathbb{R}^{N \times D_v}$. These clip features are then passed through an FC layer to reduce their dimension, denoted as $F^{1d} \in \mathbb{R}^{N \times d_v}$. Afterwards, we encode them as 2D temporal feature map $\hat{F} \in \mathbb{R}^{N \times N \times d_v}$ following 2D-TAN~\cite{zhang2020learning} with the max pooling. As for language encoder, we choose DistilBERT~\cite{sanh2019distilbert} to obtain sentence-level feature $\hat{f^{s}} \in \mathbb{R}^{d_s}$ following ~\cite{wang2022negative}. 
Finally, to estimate the matching scores of candidate moments and the query, we utilize a linear projection layer to project the textual and visual features into same dimension $d$, respectively. The final representation of sentence is $f^{s} \in \mathbb{R}^{d}$ and the features of all moments are $F \in \mathbb{R}^{N \times N \times d}$. The final matching scores are given by the cosine similarity between $f^s$ and elements of $F$.
\subsection{Semantic Alignment Group Contrastive Learning}
\begin{figure*}[t]
    \centering
    \includegraphics[width=\linewidth]{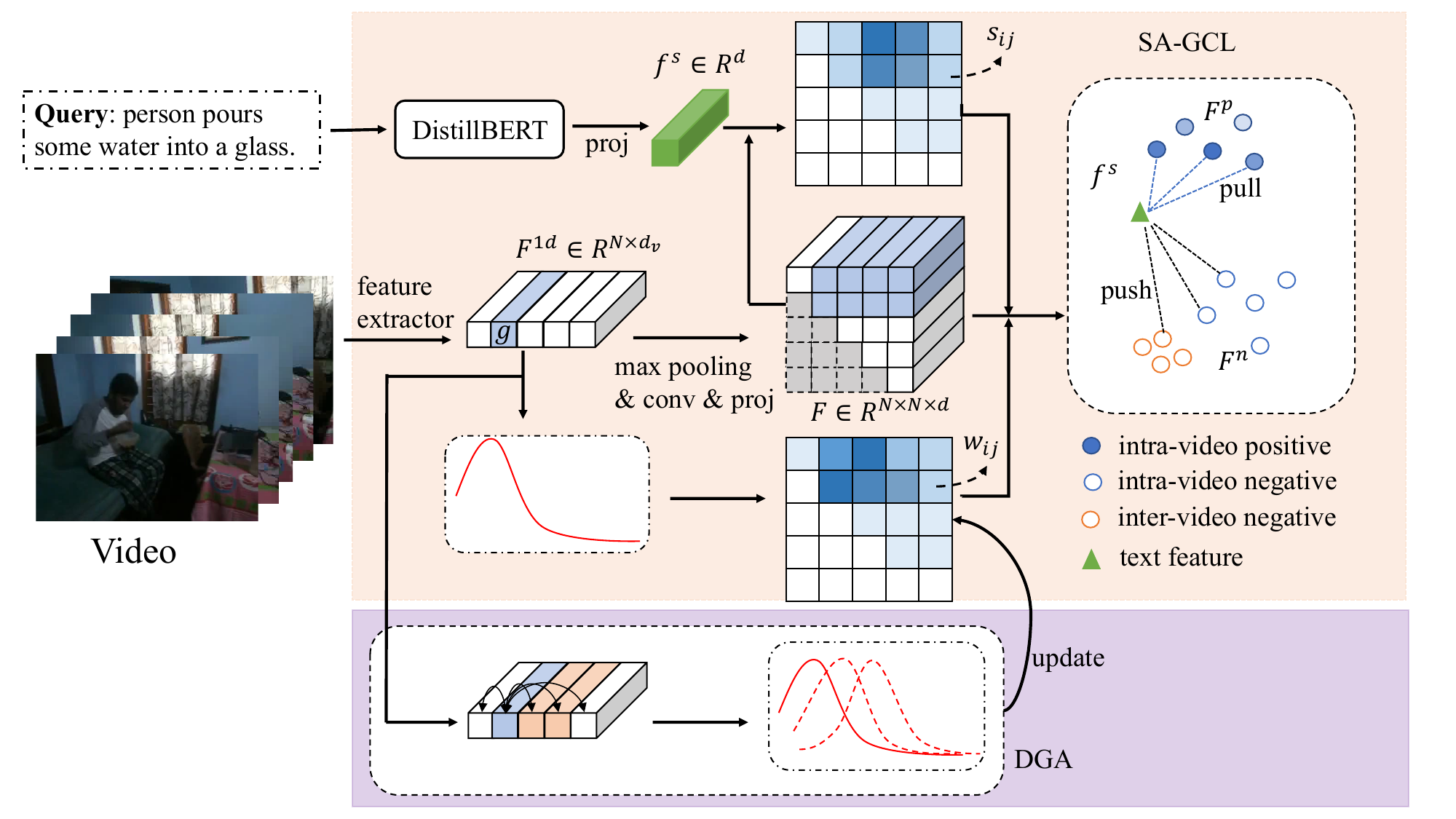}
    \caption{The overview of proposed D3G, which consists of Semantic Alignment Group Contrastive Learning (SA-GCL) and Dynamic Gaussain prior Adjustment (DGA). Note that $g$ indicates the position of glance annotation and the grids with dashed line in $F$ are invalid candidate moments. ``proj" denotes the linear projection layer. \blue{``intra/inter'' indicate the positive or negative moments sampled from same/different videos.}}
    \label{fig:framework}
\end{figure*}
\label{sec:sa_gcl}
\delete{The definition of glance annotation implies that the glance $g$ is inside the target moment. 
Intuitively, frames close to glance $g$ should have high relevance with the target moment while frames distant from glance $g$ may contain less semantic relevance with target moment. Motivated by this insight, Cui~\etal~\cite{cui2022video} propose the method ViGA, which utilizes the Gaussian distribution to model such relevance. }
\blue{In this section, we aim to mine the moment which most semantically corresponds to the query and maximize the similarity between them. To achieve this goal, we have two crucial steps. First, we generate abundant candidate moments following 2D-TAN and assign them with reliable Gaussian prior weights generated with the guidance of glance annotation. Second, we propose a semantic alignment group contrastive learning to align a group of positive moments with corresponding query sentence. }

\delete{However, ViGA is still confronted with few limitations as mentioned in Introduction.
To get rid of these limitations, we utilize the characteristic of 2D-TAN to generate candidate moments of various durations and propose a semantic alignment group contrastive loss to align a group of positive moments with corresponding query sentence.}  

To be specific, given the encoded video features $F^{1d} \in \mathbb{R}^{N \times d_v}$ and glance annotation $g$, we also utilize a Gaussian function parameterized with 
$(\mu,\sigma)$ to model the relations between frames and target moment, where the $\mu$ is determined by the glance $g$. We first scale the sequence indices $I \in \{1, 2, ..., N\}$ into domain $[-1, 1]$ by a linear transformation as follows:
\begin{equation}
    h(i) = 2 \cdotp \frac{i-1}{N-1} - 1.
\end{equation}

Given the index $i$, we can obtain corresponding Gaussian weight via Eq.~(\ref{eq:gaussian_func}).
\begin{equation}
    G(i, \mu, \sigma) = Norm( \frac{1}{\sqrt{2\pi}\sigma} exp(-\frac{(h(i)-h(\mu))^2}{2\sigma^{2}})),
    \label{eq:gaussian_func}
\end{equation}
where $\mu \in I$ and $\sigma$ is a hyperparameter, and Norm($\cdot$) is a function used to scale values into range [0, 1].

Different from ViGA~\cite{cui2022video}, we utilize the characteristic of 2D-TAN to generate a wide range of candidate moments with various durations. Given the video features $F^{1d} \in \mathbb{R}^{N \times d_v}$, we encode them into 2D feature map $F \in \mathbb{R}^{N \times N \times d}$ as shown in Figure~\ref{fig:framework}, where $F_{ij}$ denotes the feature of moment that starts at position $i$ and ends at position $j$. Note that the moment is valid only when $i \leq j$. 
We then propose a triplet-sample strategy to generate more reasonable weights for candidate moments instead of only sampling the weight at middle point as in ~\cite{cui2022video}. Specifically, for each moment with start position $i$ and end position $j$, we compute its Gaussian prior weight as follows:
\begin{equation}
    w_{ij} = \frac{1}{3} \cdotp (G(i, g, \sigma) + G(j, g, \sigma) + G(\lfloor \frac{i+j}{2} \rfloor, g, \sigma)),
    \label{eq:triplet_weight}
\end{equation}
where $g$ is glance annotation for current target moment. In this way, those moments containing target moment but having longer durations will be penalized with lower weights.


To remedy the annotation bias, we additionally introduce semantic consistency prior to calibrate the Gaussian prior weight $w_{ij}$ for each candidate moment. Given the query features $f^s \in \mathbb{R}^d$ and the features $F \in \mathbb{R}^{N \times N \times d}$ of candidate moments, we compute their semantic consistency scores via Eq.~(\ref{eq:align}).
\begin{equation}
    s_{ij} = \frac{f^s \cdot F_{ij}}{\parallel f^s \parallel \parallel F_{ij} \parallel},
    \label{eq:align}
\end{equation}
where $\parallel \cdot \parallel$ is $\mathit{l}_2$-norm. Afterwards, we rectify the Gaussian weight $w_{ij}$ with semantic consistency score $s_{ij}$ via multiplication to obtain new prior weight $p_{ij} = w_{ij} \cdotp s_{ij}$.

The objective of Temporal Sentence Grounding is to learn a cross-modal embedding space, where the query sentence feature should be well aligned with the feature of corresponding moment and far way from those of irrelevant video moments. 
Motivated by ~\cite{wang2021self,li2022siod}, we propose a Semantic Alignment Group Contrastive Learning module (SA-GCL) to gradually mine candidate moments most semantically aligned with given query sentence. To be specific, we first sample top-$k$ candidate moments from $F$ as positive keys for query $f^s$ according to the new prior $p_{ij}$, denoted as $F^p = \{ F_{ij} | 1 \leq i \leq j \leq N\} \in \mathbb{R}^{k \times d}$. Simultaneously, we sample Gaussian weights of corresponding moments denoted as $W^p = \{ w_{ij} | 1 \leq i \leq j \leq N\} \in \mathbb{R}^k$. We then gather other candidate moments which do not contain the glance $g$ from intra-video and all candidate moments from other videos within same batch as negative keys, denoted as $F^n = \{ F_{ij} | 1 \leq i \leq j \leq N\} \in \mathbb{R}^{N_n \times d}$, where $N_n$ denotes the number of negative moments. The objective of SA-GCL can be described as follows:
\begin{equation}
    \begin{aligned}
        L_{align} =& - \frac{1}{k} \sum_{z=0}^{k} W_{z}^{p} \log \frac{exp(f^s \cdot F_z^{p} / \tau)}{\textit{SUM}},  \\
        \textit{SUM} =& \sum_{z=0}^{k} exp(f^s \cdot F_z^{p} / \tau) + \sum_{z=0}^{N_n} exp(f^s \cdot F_z^{n} / \tau),
    \end{aligned}
\end{equation}
where $\tau$ is the temperature scaling factor. SA-GCL aims to maximize the similarity between the query $f^s$ and a group of corresponding positive moments $F^p$ under the joint embedding space while pushing away negative pairs. Note that different positive moments are assigned with corresponding prior weight $W_z^p$. In this way, SA-GCL effectively avoids being dominated by inaccurate moments with less similarity and 
tends to mine the candidate moments having large overlap with the target moment.

\subsection{Dynamic Gaussian prior Adjustment}
\label{sec:dga}
To further ease the annotation bias and characterize complex target moments, we propose a novel Dynamic Gaussian prior Adjustment module (DGA). Specifically, we utilize multiple Gaussian functions with different centers to model the local distributions of target moment and aggregate them to approximate the distribution of target moment. 

Given the video features $F^{1d} \in \mathbb{R}^{N \times d_v}$ and annotation glance $g$, we compute the relevance of other position $i$ with position $g$ via Eq.~(\ref{eq:rel}).
\begin{equation}
    r_{gi} = \frac{F^{1d}_g \cdot F^{1d}_i}{\parallel F^{1d}_g \parallel \parallel F^{1d}_i \parallel}. 
    \label{eq:rel}
\end{equation}
\begin{equation}
    \bar{r}_{gi} =  (1-\alpha)\bar{r}_{gi} + \alpha r_{gi}.
    \label{eq:moving_rel}
\end{equation}

To make the relevance scores more stable, we update $\bar{r}_{gi}$ with momentum factor $\alpha$ as shown in Eq.~(\ref{eq:moving_rel}), where $\bar{r}_{gi} = r_{gi}$ at first training epoch. According to the relevance $\{\bar{r}_{gi}\}$, we can mine latent local centers for target moment. Specifically, we utilize a specific threshold $T_r$ to filter the candidate positions and obtain a mask $M_g \in \{0,1\}^{N}$ for glance $g$ as follows:
\begin{equation}
    M_g^i = \begin{cases}
    1, \quad if\ \bar{r}_{gi} \geq T_{r} \\
    0, \quad otherwise
    \end{cases}
    \label{eq:rel_mask}
\end{equation}

With the mask of latent local centers, we then adjust the Gaussian prior dynamically via Eq.~(\ref{eq:dga}).
\begin{equation}
    \hat{G}(i, g, \sigma) = \frac{1}{C} \sum_{z=1}^{N} M_g^z \cdot \bar{r}_{gi} \cdot G(i, z, \sigma), 
    \label{eq:dga}
\end{equation}
where C is the summation of mask $M_g$. Afterwards, we replace the $G(i, g, \sigma)$ in Eq.~(\ref{eq:triplet_weight}) with $\hat{G}(i, g, \sigma)$, and naturally obtain dynamic Gaussian prior weight during training. Compared to ViGA, our dynamic Gaussian prior is more flexible and able to adjust the center of Gaussian distribution adaptively. Therefore, DGA further alleviates the annotation bias and provides more reliable prior weights. Besides, multiple Gaussian distributions are well suited for modeling complex target moments consisting of multiple events as shown in Figure~\ref{fig:multi_events}. DGA tends to widen the region of high Gaussian weight via self-mining neighboring frames based on the feature of glance $g$ and gradually generates the Gaussian prior weight well aligned with target moment. In this way, SA-GCL will be provided with positive moments of high quality, which eventually promotes the cross-modal semantic alignment learning and accurate localization of target moments.

\noindent \textbf{Discussion.} To clearly distinguish the differences between D3G and few similar works, we give some explanations here. As for MMN, D3G shares the same process of generating candidate moments following 2D-TAN, which is not the key contribution of our method. MMN utilizes normal one-to-one contrastive learning is no longer suitable to glance annotation. 
However, D3G instead adopts a suitable sample strategy and corresponding adapted group contrastive learning,
which is key component to unleash the potential of glance annotations. As for CPL, we also know that it utilizes multiple Gaussian distribution to describe positive moments. However, it actually select one most
matched positive moment guided by the loss of masked language reconstruction for contrastive learning, while D3G utilizes multiple Gaussian functions to adaptively model
complex queries consisting of multiple events and samples a group of positive moments for contrastive learning. 
\section{Experiments}
In order to validate the effectiveness of the proposed D3G, we conduct extensive experiments on three publicly available datasets: Charades-STA~\cite{gao2017tall}, TACoS~\cite{gao2017tall} and ActivityNet Captions~\cite{krishna2017dense}.
\subsection{Datasets}
\noindent \textbf{Charades-STA} is built on dataset Charades~\cite{sigurdsson2016hollywood} for temporal sentence grounding. It contains 12,408 and 3,720 moment-sentence pairs for training and testing.

\noindent \textbf{TACoS} consists of 127 videos selected from the MPII Cooking Composite Activities video corpus~\cite{rohrbach2012script}. We follow the standard split from~\cite{gao2017tall}, which contains 10,146, 4,589 and 4,083 moment-sentence pairs for training, validation and testing, respectively. We report the evaluation result on the test set for fair comparison.

\noindent \textbf{ActivityNet Captions} is originally designed for video captioning and recently introduced into temporal sentence grounding. It contains 37,417, 17,505 and 17,031
moment-sentence pairs for training, validation and testing, respectively. We report the evaluation result 
 following ~\cite{zhang2020learning, wang2022negative}.

Specially, we adopt the glance annotation released by~\cite{cui2022video} for training set, where the temporal boundary is replaced with the timestamp $g$ uniformly sampled within the original temporal boundary.
\subsection{Evaluation Metric and Implementation Details}
\noindent \textbf{Evaluation Metric.} Following previous works~\cite{gao2017tall,zhang2020learning}, we evaluate our model with metric `R@$n$,IoU=$m$', which means the percentage of at least one of the top-$n$ results having Intersection over Union (IoU) larger than $m$. Specifically, we report the results with $m \in \{0.5, 0.7 \}$ for Charades-STA, $m \in \{0.3, 0.5, 0.7 \}$ for TACoS and ActivityNet Captions, and $n \in \{1, 5\}$ for all datasets.

\noindent \textbf{Implementation Details.} In this work, our main framework is extended from MMN~\cite{wang2022negative} and most of experiment settings keep the same. For fair comparison, following~\cite{wang2022negative}, we adopt off-the-shelf video features for all datasets (VGG feature for Charades and C3D feature for TACoS and ActivityNet Captions).
Specifically, the dimension of joint feature space $d$ is set to 256 and $\tau$ is set to 0.1. In SA-GCL, we set the $k$ as 10, 20 and 20 for Charades, TACoS and ActivityNet Captions, respectively. The $\sigma$ in Eq.~(\ref{eq:gaussian_func}) is set to 0.3, 0.2 and 0.6 for Charades, TACoS and ActivityNet Captions. In DGA, $T_r$ and $\alpha$ is set as 0.9 and 0.7, respectively.
\subsection{Comparisons with the State-Of-The-Art}
In order to provide comprehensive analysis, we compare the proposed D3G with both fully/weakly/glance supervised methods. As shown in Table~\ref{tab:sota_charades}, Table~\ref{tab:sota_tacos} and Table~\ref{tab:sota_activitynet}, D3G achieves highly competitive results on three datasets under glance supervision, and achieves comparable performance compared with fully supervised methods. Note that we highlight the best value for each setting respectively. Based on the experimental results, we can draw the following conclusions:

(1) Glance annotation provides more potential to achieve better performance for temporal sentence grounding with lower annotation cost. Although it is not entirely fair to directly compare D3G with other weakly supervised methods due to introducing extra supervision, D3G significantly exceeds most of weakly supervised methods by a large margin with trivial increment of annotation cost. 
Since PS-VTG and PFU adopt more robust I3D feature, they obviously outperform D3G on Charades-STA. However, D3G instead is superior to PS-VTG on more challenging TACoS with same features.
Besides, weak supervised methods are often not tested on TACoS, where the videos are very long and contain a large number of target moments. However, D3G obtains promising performance and outperforms ViGA by a large margin on TACoS as shown in Table~\ref{tab:sota_tacos}.

\begin{table}[ht]
\setlength\tabcolsep{5pt}
    \begin{center}
    \begin{tabular}{l|c c|c c}
        \toprule
        \multirow{2}{*}{Method} & \multicolumn{2}{c|}{R@1} & \multicolumn{2}{c}{R@5} \\
        &  IoU=0.5 & IoU=0.7 &  IoU=0.5 & IoU=0.7  \\
        \midrule
        MAN~\cite{zhang2019man} & 41.21 & 20.54 & 83.21 & 51.85 \\
        2D-TAN~\cite{zhang2020learning} & 39.70 & 23.31 & 80.32 & 51.26 \\ 
        SSCS~\cite{ding2021support} & 43.15 & 25.54 & \textbf{84.26} & 54.17 \\
        MMN~\cite{wang2022negative} & \textbf{47.31} & \textbf{27.28} & 83.74 & \textbf{58.41} \\
        \midrule
        CRM~\cite{huang2021cross} & 34.76 & 16.37 & - & - \\ 
        CNM~\cite{zheng2022weakly1} & 35.43 & 15.45 & - & - \\ 
        LCNet~\cite{yang2021local} & \textbf{39.19} & \textbf{18.87} & \textbf{80.56} & \textbf{45.24} \\ 
        CPL$^\dagger$~\cite{zheng2022weakly} & 32.27 & 14.22 & 78.34 & 43.45  \\ 
        \midrule
        PS-VTG$^\ddagger$~\cite{xu2022point} & 39.22 & \textcolor{blue}{20.17} & - & - \\ 
        PFU$^\ddagger$~\cite{ju2023constraint} & \textbf{54.66} & \textbf{28.34} & - & - \\ 
        VIGA$^*$~\cite{cui2022video} & 36.56 & 16.10 & 48.90 & 25.86 \\
        \textbf{D3G} & \textcolor{blue}{41.64} & 19.60 & \textbf{79.25} & \textbf{49.30} \\
        \bottomrule  
    \end{tabular}
    \end{center}
    \caption{Performance comparison on Charades-STA under different supervision settings.Top:full supervision, Middle: weak supervision, Bottom:glance supervision. $\dagger$we reproduce the results with official code and VGG features for fair comparison. $^*$we reproduce the results with official code for results at R@5.$^\ddagger$ indicates the method utilizes I3D features.}
    \label{tab:sota_charades}
\end{table}


\begin{table}[ht] 
    \small
    \setlength\tabcolsep{3pt}
    \begin{center}
    \renewcommand\arraystretch{1} 
    \begin{tabular}{l|c|c|c|c|c|c} 
        \toprule
        \multirow{2}{*}{Method} & \multicolumn{3}{c|}{R@1} & \multicolumn{3}{c}{R@5} \\
        & {\scriptsize IoU=0.3} & {\scriptsize IoU=0.5} & {\scriptsize IoU=0.7} & {\scriptsize IoU=0.3} & {\scriptsize IoU=0.5} & {\scriptsize IoU=0.7}  \\
        \midrule 
        CTRL~\cite{gao2017tall} & 18.32 & 13.30 & - & 36.69 & 25.42 & - \\
        2D-TAN~\cite{zhang2020learning} & 37.29 & 25.32 & - & 57.81 & 24.04 & - \\
        SSCS~\cite{ding2021support} & 41.33 & 29.56 & - & 60.65 & 48.01 & - \\
        MMN~\cite{wang2022negative} & 38.57 & 27.24 & - & 65.31 & 50.69 & - \\
        MAT~\cite{zhang2021multi} & \textbf{48.79} & \textbf{37.57} & - & \textbf{67.63} & \textbf{57.91} & - \\
        \midrule
        VIGA$^*$~\cite{cui2022video} & 20.82 & 9.52 & 3.10 & \textcolor{blue}{27.92} & \textcolor{blue}{15.35} & \textcolor{blue}{6.10} \\
        PS-VTG~\cite{xu2022point} & \textcolor{blue}{23.64} & \textcolor{blue}{10.00} & \textcolor{blue}{3.35} & - & - & - \\
        \textbf{D3G} & \textbf{27.27} & \textbf{12.67} & \textbf{4.70} & \textbf{54.61} & \textbf{31.34} & \textbf{12.35} \\
        \bottomrule  
    \end{tabular}
    \end{center}
    \caption{Performance comparison on TACoS under different supervision settings.Top:full supervision, Bottom:glance supervision. $^*$we reproduce the results with official code for results at R@5.}
    \label{tab:sota_tacos}
    \vspace{-0.5cm}
\end{table} 
(2) D3G effectively exploits the information provided by glance annotation and mines more moments of high quality for training compared with ViGA. Due to the limitations of fixed scale Gaussian function and fixed sliding window, ViGA fails to mine accurate candidate moments to learn a well-aligned joint embedding space. Instead, D3G 
generates a wide range of candidate moments and samples a group of reliable candidate moments for group contrastive learning. 
Compared to ViGA, D3G achieves obvious gains 5.08\% and 3.5\% at R@1 IoU=0.5 and R@1 IoU=0.7 on Charades-STA, respectively. Specially, significant improvements are obtained at R@5 on three datasets. 

(3) D3G substantially narrows the performance gap between weakly/glance supervised methods and fully supervised methods. Specifically, D3G already surpasses previous method (\eg, CTRL) on both TACoS and ActivityNet Captions. Undeniably, there are still non-negligible margin compared to the state-of-the-art fully supervised methods (\eg, MMN). Note that D3G is very concise and not embedded with auxiliary module (\eg, MLM used in~\cite{zheng2022weakly}). D3G still can be enhanced with some complementary modules.  
\newcommand\viswidth{0.3}
\begin{figure*}
    \centering
    \includegraphics[width=\linewidth, trim=0 645 0 10, clip]{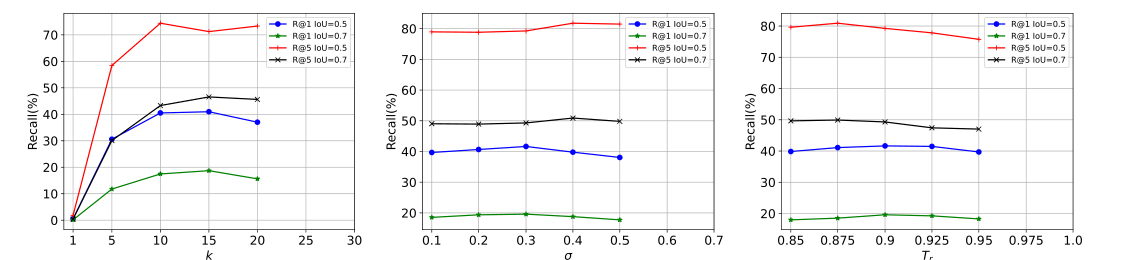} \\
    \begin{minipage}{\viswidth\linewidth}
        \centering
        \figtext{(a) Effect of top-$k$.}
    \end{minipage}
    \begin{minipage}{\viswidth\linewidth}
        \centering
        \figtext{(b) Effect of $\sigma$.}
    \end{minipage}
    \begin{minipage}{\viswidth\linewidth}
        \centering
        \figtext{(c) Effect of $T_r$}
    \end{minipage}
    \caption{Effect of different hyper-parameters on Charades-STA dataset.}
    \label{fig:ab_hp_merge}
\end{figure*}
\begin{table}[t] \small
    \setlength\tabcolsep{3pt}
    \begin{center}
    \begin{tabular}{l|c|c|c|c|c|c}
        \toprule
        \multirow{2}{*}{Method} & \multicolumn{3}{c|}{R@1} & \multicolumn{3}{c}{R@5} \\
        & {\scriptsize IoU=0.3} & {\scriptsize IoU=0.5} & {\scriptsize IoU=0.7} & {\scriptsize IoU=0.3} & {\scriptsize IoU=0.5} & {\scriptsize IoU=0.7}  \\
        \midrule
        CTRL~\cite{gao2017tall} & 47.43 & 29.01 & 10.34 & 75.32 & 59.17 & 37.54 \\
        2D-TAN~\cite{zhang2020learning} & 59.46 & 44.51 & 26.54 & 85.53 & 77.13 & 61.96 \\
        LGI~\cite{mun2020local} & 58.52 & 41.51 & 23.07 & - & - & -  \\
        SSCS~\cite{ding2021support} & 61.35 & 46.67 & 27.56 & 86.89 & 78.37 & 63.78 \\
        MMN~\cite{wang2022negative} & \textbf{65.05} & \textbf{48.59} & 29.26 & \textbf{87.25} & \textbf{79.50} & \textbf{64.76} \\
        MAT~\cite{zhang2021multi} & - & 48.02 & \textbf{31.78} & - & 78.02 & 63.18 \\
        \midrule
        CRM~\cite{huang2021cross} & 55.26 & 32.19 & - & - & - & - \\
        CNM~\cite{zheng2022weakly1} & \textbf{55.68} & \textbf{33.33} & - & - & - & - \\
        LCNet~\cite{yang2021local} & 48.49 & 26.33 & - & \textbf{82.51} & \textbf{62.66} & - \\
        CPL~\cite{zheng2022weakly} & 53.67 & 31.24 & - & 63.05 & 43.14 & -  \\
        \midrule
        VIGA$^*$~\cite{cui2022video} & \textbf{59.78} & 35.39 & 16.25 & 72.19 & 53.19 & 32.69 \\
        PS-VTG~\cite{xu2022point} & \textcolor{blue}{59.71} & \textbf{39.59} &  \textbf{21.98} & - & - & - \\
        PFU~\cite{ju2023constraint} & 59.63 & 36.35 & 16.61 & - & - & - \\ 
        \textbf{D3G} & 58.25 & \textcolor{blue}{36.68} & \textcolor{blue}{18.54} & \textbf{87.84} & \textbf{74.21} & \textbf{52.47}  \\
        \bottomrule  
    \end{tabular}
    \end{center}
    \caption{Performance comparison on ActivityNet Captions under different supervision settings.Top:full supervision, Middle: weak supervision, Bottom:glance supervision. $^*$we reproduce the results with official code for results at R@5.}
    \label{tab:sota_activitynet}
    \vspace{-0.5cm}
\end{table}
\vspace{-0.5cm}
\subsection{Ablation Study} 
To validate the effectiveness of different components of the proposed D3G and investigate the impact of hyper-parameters, we perform ablation studies on Charades-STA.

\noindent \textbf{Effectiveness of SA-GCL and DGA.} \delete{We first investigate the importance of Semantic Alignment Group Contrastive Learning module and Dynamic Gaussian prior Adjustment module. }Since $L_{align}$ is the only loss of D3G, to validate the effectiveness of SA-GCL, we need to simplify the SA-GCL module as a baseline. Specifically, we only sample the top-$1$ positive moment to compute the normal contrastive loss (degraded to simplified MMN) as shown in the first row of Table~\ref{tab:ab_d3g}. However, the top-$1$ moment tends to be the shortest moment and has small overlap with target moment, which is decided by the intrinsic characteristic of 2D-TAN. Therefore, the performance of baseline is undoubtedly very poor, which demonstrates that the main improvement of D3G is not brought by the backbone of MMN. This phenomenon then encourages us to sample a group of positive moments in SA-GCL. 
With full SA-GCL, the model obtains notable performance gains. Moreover, we introduce the DGA to alleviate annotation bias and model some complex target moments consisting of multiple events. After equipped with DGA, D3G and simplified D3G achieve obvious performance improvement. 

\noindent \textbf{Impact of Sampling Strategy.} In SA-GCL, sampling a group of reliable positive moments is of great importance. We investigate the impacts of two priors: Gaussian weight and semantic consistency, respectively. As shown in the first row of Table~\ref{tab:ab_sample}, we sample top-$k$ positive moments according to the Gaussian prior weight. An alternative scheme is that we sample top-$k$ positive moments according to the semantic consistency scores between candidate moments and query sentence. However, both of them obtain sub-optimal performance. This is because Gaussian prior weight is not always reliable due to the annotation bias and semantic consistence scores are highly dependent on the stability of features. Therefore, we finally fuse these two priors to obtain relatively reliable prior. As shown in the third row of Table~\ref{tab:ab_sample}, obvious performance gains are obtained after both of them are utilized, which demonstrates that these two priors indeed complement each others.

\begin{table}[t]
\setlength\tabcolsep{3.5pt}
    \begin{center}
    \begin{tabular}{c c|c c|c c}
        \toprule
        \multicolumn{2}{c|}{Module} & \multicolumn{2}{c|}{R@1} & \multicolumn{2}{c}{R@5} \\
        SA-GCL & DGA &  IoU=0.5 & IoU=0.7 &  IoU=0.5 & IoU=0.7  \\
        \midrule
        \checkmark$^\dagger$& & 5.08 & 0.81 & 14.78 & 3.36 \\
        \checkmark$^\dagger$& \checkmark & 13.92 & 3.31 & 33.55 & 11.77 \\
        \checkmark & & 40.51 & 16.10 & 74.41 & 43.31 \\
        \checkmark & \checkmark & 41.64 & 19.60 & 79.25 & 49.30 \\
        \bottomrule  
    \end{tabular}
    \end{center}
    \caption{Effectiveness of SA-GCL and DGA in D3G on Charades-STA. $\checkmark^\dagger$ denotes an simplified implementation of SA-GCL.}
    \label{tab:ab_d3g}
\end{table}

\begin{table}[t]
\setlength\tabcolsep{6pt}
    \begin{center}
    \begin{tabular}{c c|c c|c c}
        \toprule
        \multicolumn{2}{c|}{Types} & \multicolumn{2}{c|}{R@1} & \multicolumn{2}{c}{R@5} \\
        GW & SC &  IoU=0.5 & IoU=0.7 &  IoU=0.5 & IoU=0.7  \\
        \midrule
        \checkmark & & 38.09 & 16.10 & 66.53 & 36.51 \\
        & \checkmark & 25.67 & 9.57 & 65.43 & 38.52 \\
        \checkmark & \checkmark & 40.51 & 16.10 & 74.41 & 43.31 \\
        \bottomrule  
    \end{tabular}
    \end{center}
    \caption{Impact of different strategies used to sample positive moments for SA-GCL on Charades-STA. GW: Gaussian weight, SC: semantic consistency.}
    \label{tab:ab_sample}
     
\end{table}

\noindent \textbf{Effect of different hyper-parameters.} As shown in Figure~\ref{fig:ab_hp_merge}, we investigate three critical hyperparameters in D3G. As verified in Table~\ref{tab:ab_d3g}, sampling enough latent positive moments is beneficial to mining target moment for training. As shown in Figure~\ref{fig:ab_hp_merge}~(a), the performance gains increase obviously as the $k$ increases. However, it begins to decrease after the $k$ reaches a specific value. We argue that selecting excessive positive moments tends to incorporate some false positive moments and therefore degrades the performance. We finally set the $k$ to 10 for Charades-STA, which balances well the performance and computational cost. As for hyperparameter $\sigma$, it essentially decides the width of Gaussian distribution. A larger $\sigma$ can well characterize the target moment of longer duration and vice versa. We vary the $\sigma$ from 0.1 to 0.5, and observe that value 0.3 is relatively suitable for the Charades-STA dataset. As for hyperparameter $T_r$ in Eq.~(\ref{eq:rel_mask}), it controls the degree of dynamic Gaussian prior adjustment. We conduct experiments with relevance thresholds around 0.9. A small threshold tends to introduce interference while a large threshold fails to find the neighbor frames with consistent semantic. As shown in Figure~\ref{fig:ab_hp_merge} (c), the moderate threshold 0.9 relatively balances the aforementioned dilemma. 
\begin{table}[t]
\setlength\tabcolsep{3pt}
    \small
    \begin{center}
    \begin{tabular}{l|c c|c c}
        \toprule
        \multirow{2}{*}{Method} & \multicolumn{2}{c|}{R@1} & \multicolumn{2}{c}{R@5} \\
        &  IoU=0.5 & IoU=0.7 &  IoU=0.5 & IoU=0.7  \\
        \midrule
        ViGA & 36.56 & 16.10 & 48.90 & 25.86 \\
        D3G & 41.64 & 19.60 & 79.25 & 49.30 \\
        ViGA$^+$ & 33.66$_{(-2.90)}$ & 14.65$_{(-1.45)}$ & 47.45$_{(-1.45)}$ & 25.51$_{(-0.35)}$ \\
        D3G$^+$ & 40.19$_{(-1.45)}$ & 19.62$_{(+0.02)}$ & 78.90$_{(-0.35)}$ & 49.41$_{(+0.11)}$ \\ 
        \bottomrule  
    \end{tabular}
    \end{center}
    \caption{Performance comparison on Charades-STA with extreme glance annotation. $^+$ indicates according method is trained with extreme glance annotations.}
    \label{tab:ab_hard}
    \vspace{-0.5cm}
\end{table}


\noindent \textbf{Tolerance to Extreme Glance Annotation.} In order to verify the ability of addressing extreme glance annotation, we first generate extreme glance annotation, where only the positions near the start/end timestamps will be sampled as glance $g$. As shown in Table~\ref{tab:ab_hard}, both ViGA$^+$ and D3G$^+$ are confronted with the performance degradation at some metrics(\eg, R@1 IoU=0.5). However, the performances of D3G are relatively stable compared to ViGA, which demonstrate that D3G indeed is able to alleviate annotation bias.

\subsection{Qualitative Analysis}
\begin{figure}
    \centering
    \includegraphics[width=\linewidth]{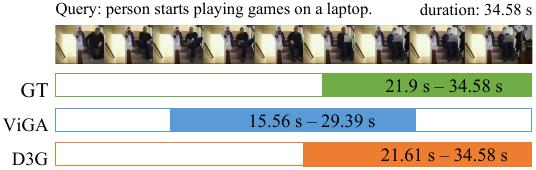} \\
    \begin{minipage}{\linewidth}
        \centering
        \figtext{(a)}
    \end{minipage}
    \includegraphics[width=\linewidth]{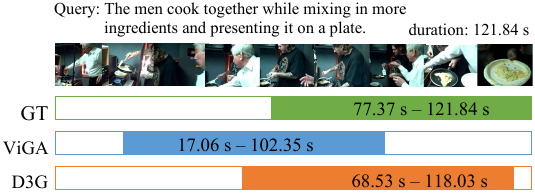} \\
    \begin{minipage}{\linewidth}
        \centering
        \figtext{(b)}
    \end{minipage}
    \caption{Qualitative examples of top-1 predictions. (a) and (b) is from the Charades-STA dataset and the ActivityNet Captions dataset, respectively. GT indicates the ground truth temporal boundary.}
    \label{fig:visualize}
    \vspace{-0.5cm}
\end{figure}
To clearly reveal the effectiveness of our method, we visualize some qualitative examples from the test split of Charades-STA dataset and ActivityNet Captions dataset.  As shown in Figure~\ref{fig:visualize}, the proposed D3G achieves more accurate localization of target moment compared to ViGA. Specifically, ViGA cannot well align the visual content and  semantic information and tend to be disturbed by irrelevant content, which may be caused by the annotation bias. Instead, D3G utilizes SA-GCL and DGA to alleviate the annotation bias, which enables D3G to well align the query with the corresponding moment. Moreover, the DGA adopts multiple Gaussian functions to model target moment, which is beneficial to representing the complete distribution of complex moments consisting of multiple events. As shown in Figure~\ref{fig:visualize} (b), D3G still effectively localizes the complex moments while ViGA misses the last events ``represent it on a plate". More qualitative examples will be provided in Supplementary Materials.  
\section{Conclusion}
In this study, we investigate a recently proposed task, Temporal Sentence Grounding with Glance Annotation. Under this setting, we propose a \textbf{D}ynamic \textbf{G}aussian prior based \textbf{G}rounding framework with \textbf{G}lance annotation, termed D3G. Specifically, D3G consists of a Semantic Alignment Group Contrastive Learning module (SA-GCL) and a Dynamic Gaussian prior Adjustment module (DGA). SA-GCL aims to mine a wide range of positive moments and align the positive sentence-moment pairs in the joint embedding space. DGA effectively alleviates the annotation bias and models complex query consisting of multiple events via dynamically adjusting the Gaussian prior with multiple Gaussian functions, promoting the precision of localization. Extensive experiments show that D3G significantly narrows the performance gap between fully supervised methods and glance supervised methods. 
Without excessive interaction of visual-language, D3G provides a concise framework and a fresh insight to the challenging temporal sentence grounding under low-cost glance annotation.

\noindent \textbf{Limitations.} Although D3G achieves promising improvements with glance annotations, it still has some limitations. In this paper, the DAG adjusts Gaussian prior via the combination of multiple fixed scale Gaussian functions. It fails to scale down the Gaussian distribution to fit the small moments. It is expected to explore dynamic learnable Gaussian functions to model moment of arbitrary duration in future work. Besides, the sampling strategy for SA-GCL is still not enough flexible to sample accurate positive moments.

{\small
\bibliographystyle{ieee_fullname}
\bibliography{egbib}
}
\clearpage
\appendix
\noindent{\Large \textbf{Appendix}}
\section{Effectiveness of SA-GCL and DGA }
To further analyze the effectiveness of SA-GCL and DGA, we provide more detailed experimental results on ActivityNet Captions and TACoS datasets as shown in Table~\ref{tab:anet_ab_d3g} and Table~\ref{tab:tacos_ab_d3g}. Following the main manuscript, we regard the simplified implementation of SA-GCL as a baseline. After being equipped with the complete SA-GCL, our model achieves significant improvements on both ActivityNet Captions and TACoS. This phenomenon demonstrates that sampling enough positive moments for contrastive learning is of great importance. Additionally, we further incorporate the DGA module for alleviating the annotation bias and modeling complex target moments. Since the ActivityNet Captions dataset has a large number of complex query sentences consisting of multiple events, D3G obtains notable performance gains on ActivityNet Captions(\eg 9.03\% at R@5 IoU=0.7). However, TACoS is still challenging for D3G due to the dense distributions of target moments.  

\begin{table}[h]
\setlength\tabcolsep{4pt}
    \begin{center}
    \begin{tabular}{c c|c c|c c}
        \toprule
        \multicolumn{2}{c|}{Module} & \multicolumn{2}{c|}{R@1} & \multicolumn{2}{c}{R@5} \\
        SA-GCL & DGA &  IoU=0.5 & IoU=0.7 &  IoU=0.5 & IoU=0.7  \\
        \midrule
        \checkmark$^\dagger$& & 0.83 & 0.28 & 1.78 & 0.58 \\
        \checkmark & & 32.65 & 16.00 & 65.48 & 43.44 \\
        \checkmark & \checkmark & 36.68 & 18.54 & 74.21 & 52.47 \\
        \bottomrule  
    \end{tabular}
    \end{center}
    \caption{Effectiveness of SA-GCL and DAG in D3G on ActivityNet Captions. $\checkmark^\dagger$ denotes an simplified implementation of SA-GCL.}
    \label{tab:anet_ab_d3g}
    \vspace{-0.3cm}
\end{table}
\begin{table}[h]
\setlength\tabcolsep{4pt}
    \begin{center}
    \begin{tabular}{c c|c c|c c}
        \toprule
        \multicolumn{2}{c|}{Module} & \multicolumn{2}{c|}{R@1} & \multicolumn{2}{c}{R@5} \\
        SA-GCL & DGA &  IoU=0.5 & IoU=0.7 &  IoU=0.5 & IoU=0.7  \\
        \midrule
        \checkmark$^\dagger$& & 2.97 & 0.37 & 5.40 & 1.10 \\
        \checkmark & & 11.95 & 4.20 & 29.07 & 10.30 \\
        \checkmark & \checkmark & 12.67 & 4.70 & 31.34 & 12.35 \\
        \bottomrule  
    \end{tabular}  
    \end{center}
    \caption{Effectiveness of SA-GCL and DAG in D3G on TACoS. $\checkmark^\dagger$ denotes an simplified implementation of SA-GCL.}
    \label{tab:tacos_ab_d3g}
\end{table}
\section{Effect of different hyper-parameters}
\begin{figure}
    \centering
    \includegraphics[width=\linewidth, trim=10 590 280 20, clip]{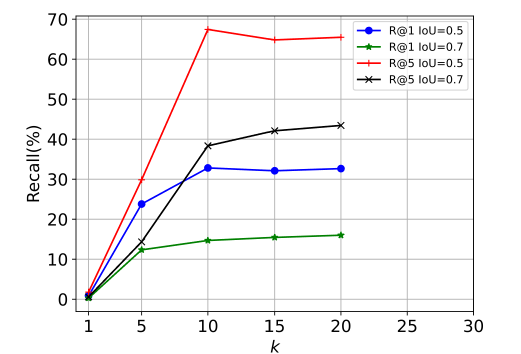} \\
    \includegraphics[width=\linewidth, trim=10 580 280 15, clip]{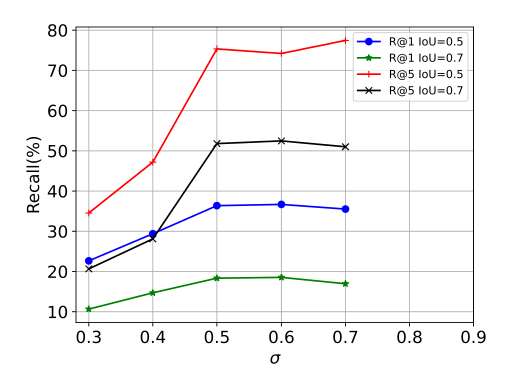} \\
    \caption{Effect of top-$k$ and $\sigma$ on ActivityNet Captions dataset.}
    \label{fig:sup_anet_ab}
\end{figure}
\begin{figure}
    \centering
    \includegraphics[width=\linewidth, trim=10 590 280 20, clip]{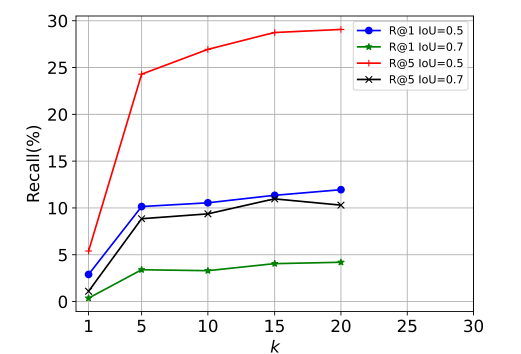} \\
    \includegraphics[width=\linewidth, trim=10 580 280 20, clip]{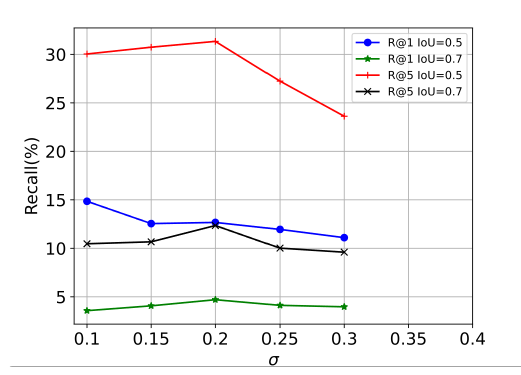} \\
    \caption{Effect of top-$k$ and $\sigma$ on TACoS dataset.}
    \label{fig:sup_tacos_ab}
\end{figure}
In this section, we investigate the effect of two critical hyperparameters on ActivityNet Captions and TACoS datasets. As shown in Figure~\ref{fig:sup_anet_ab} and Figure~\ref{fig:sup_tacos_ab}, we report the changes in performance at four metrics. As for top-$k$, the performance increases dramatically as the $k$ increases. However, the performance gradually achieves saturation after the $k$ reaches 15. We finally select $k=20$ for both ActivityNet Captions and TACoS. As for $\sigma$, the ActivityNet Captions dataset tends to prefer large values while small values are more suitable for the TACoS dataset. This is because the former contains a large number of long target moments while the latter contains numerous short target moments. As shown in Figure~\ref{fig:sup_anet_ab} and Figure~\ref{fig:sup_tacos_ab}, we eventually select $\sigma=0.6$ and  $\sigma=0.2$ for ActivityNet Captions and TACoS for optimal performance, respectively.

\section{Qualitative Analysis}
\begin{figure*}
    \centering
    \includegraphics[width=\linewidth, trim=0 125 0 0, clip]{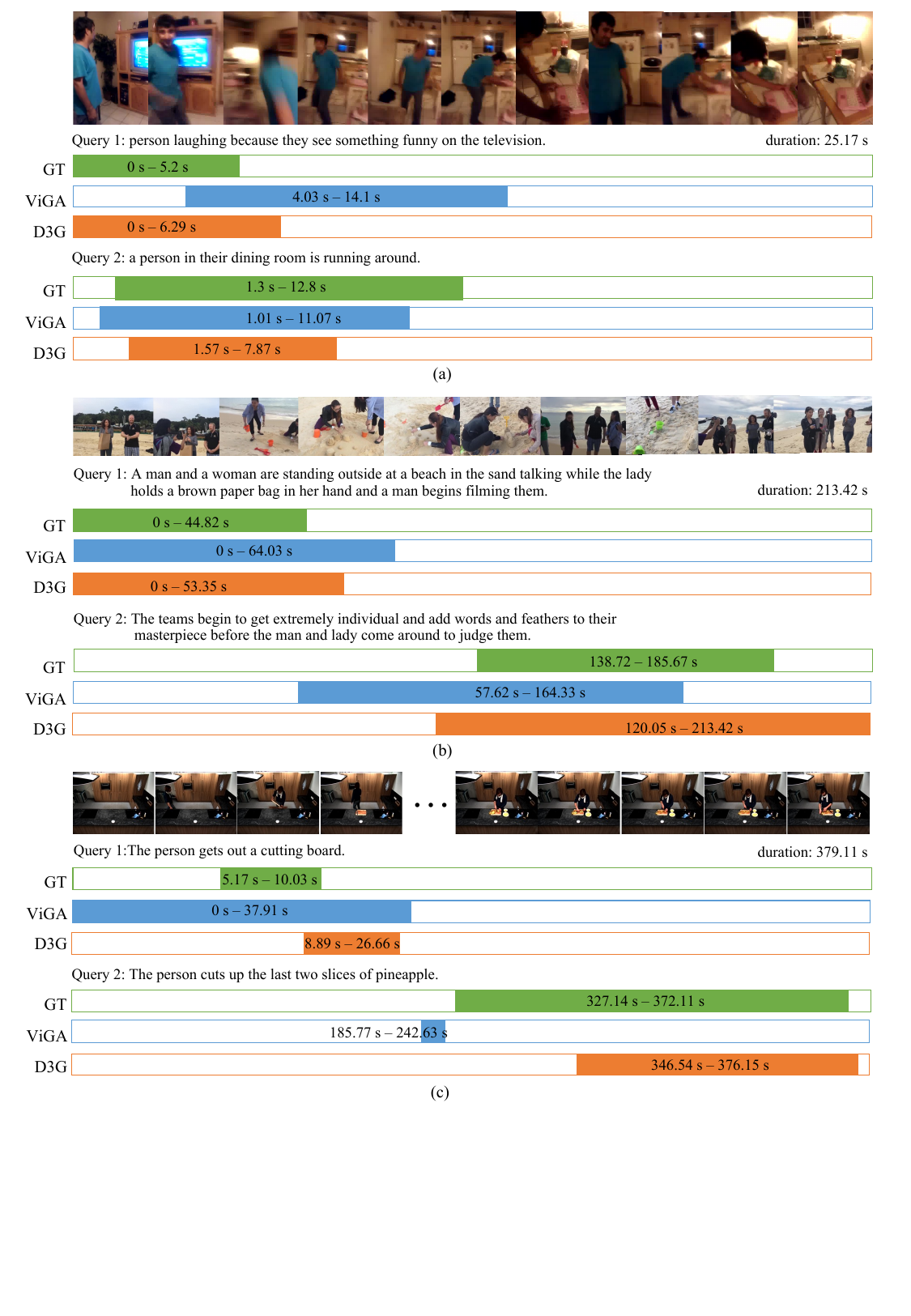}
    \caption{Qualitative examples of top-1 predictions. (a), (b) and (c) is from the Charades-STA dataset, the ActivityNet Captions and the TACoS dataset, respectively. GT indicates the ground truth temporal boundary.}
    \label{fig:sup_visualize}
\end{figure*}

In this section, we provide more qualitative examples from the test split of the Charades-STA dataset, ActivityNet Captions dataset, and TACoS dataset. For each video, we select two queries for analysis. As shown in Figure~\ref{fig:sup_visualize} (a), D3G locates the target moment accurately while ViGA ignores the reason at the front of the target moment, given Query 1. However, D3G is inferior to ViGA in some cases such as Query 2. As for complex queries in ActivityNet Captions, D3G still localizes a moment with a large overlap with the target moment. Since sentence-level features may lose some information about specific events, D3G cannot perceive accurate boundaries for some complex queries, such as Figure~\ref{fig:sup_visualize} (b) Query 2. It is expected to explore event-level features for queries consisting of multiple events in the future. TACoS is the most challenging dataset, where the videos have long durations and contain a large number of moment-sentence pairs.  As shown in Figure~\ref{fig:sup_visualize} (c), we observe that D3G fails to locate a simple query of short duration from the long video, given Query 1. However, D3G accurately locates the target moment of long duration given Query 2. Note that D3G well attends to the number ``the last two'' of the query while ViGA fails to attend to such information and locates irrelevant moments. As observed in Figure~\ref{fig:sup_visualize}, D3G is superior to ViGA, which is consistent with the experimental results in the main manuscript. However, D3G still has some limitations and needs to be improved in the future.  

\end{document}